\documentclass[10pt,conference]{IEEEtran}

\usepackage{cite}
\usepackage{amsmath,amssymb,amsfonts}
\usepackage[cmintegrals]{newtxmath}
\usepackage{bm}

\usepackage{graphicx}
\usepackage{textcomp}
\usepackage{graphicx}    
\usepackage{subcaption}  
\usepackage{caption}     
\usepackage{xcolor}
\usepackage[utf8]{inputenc} 
\usepackage{alphabeta} 
\usepackage[ruled,vlined,algo2e]{algorithm2e}
\usepackage[hidelinks]{hyperref}
\usepackage{diagbox}
\usepackage{threeparttable}  
\usepackage{academicons} 
\usepackage{booktabs}   
\usepackage{multirow}   
\usepackage{array} 
\newcommand{\orcidicon}[1]{\href{https://orcid.org/#1}{\raisebox{0.5ex}{\includegraphics[height=1.5ex]{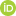}}}}

\newtheorem{definition}{Definition}

\usepackage{algorithm,algpseudocode}

\def\BibTeX{{\rm B\kern-.05em{\sc i\kern-.025em b}\kern-.08em
    T\kern-.1667em\lower.7ex\hbox{E}\kern-.125emX}}

\begin{document}

\title{Transformer-based Multivariate Time Series Anomaly Localization}

\author{
	\IEEEauthorblockN{
		Charalampos Shimillas\textsuperscript{1,2} \orcidicon{0009-0000-8878-9664}, 
		Kleanthis Malialis\textsuperscript{1} \orcidicon{0000-0003-3432-7434}, 
		Konstantinos Fokianos\textsuperscript{3} \orcidicon{0000-0002-0051-711X}, 
		Marios M. Polycarpou\textsuperscript{1,2} \orcidicon{0000-0001-6495-9171}
	}

	\IEEEauthorblockA{
		\textsuperscript{1}\textit{KIOS Research and Innovation Center of Excellence, Nicosia, Cyprus} \\
		\textsuperscript{2}\textit{Department of Electrical and Computer Engineering, University of Cyprus, Nicosia, Cyprus} \\
		\textsuperscript{3}\textit{Department of Mathematics and Statistics, University of Cyprus, Nicosia, Cyprus} \\
		\{shimillas.charalampos, malialis.kleanthis, fokianos.konstantinos, mpolycar\}@ucy.ac.cy 
	}
}

\maketitle

\begin{abstract}
With the growing complexity of Cyber-Physical Systems (CPS) and the integration of Internet of Things (IoT), the use of sensors for online monitoring generates large volume of multivariate time series (MTS) data. Consequently, the need for robust anomaly diagnosis in MTS is paramount to maintaining system reliability and safety. While significant advancements have been made in anomaly detection, localization remains a largely underexplored area, though crucial for intelligent decision-making. This paper introduces a novel transformer-based model for unsupervised anomaly diagnosis in MTS, with a focus on improving localization performance, through an in-depth analysis of the self-attention mechanism's learning behavior under both normal and anomalous conditions. We formulate the anomaly localization problem as a three-stage process: time-step, window, and segment-based. This leads to the development of the Space-Time Anomaly Score (STAS), a new metric inspired by the connection between transformer latent representations and space-time statistical models. STAS is designed to capture individual anomaly behaviors and inter-series dependencies, delivering enhanced localization performance. Additionally, the Statistical Feature Anomaly Score (SFAS) complements STAS by analyzing statistical features around anomalies, with their combination helping to reduce false alarms. Experiments on real world and synthetic datasets illustrate the model’s superiority over state-of-the-art methods in both detection and localization tasks.
\end{abstract}


\section{Introduction}
\footnotetext{This paper was supported by the European Union’s Horizon Europe research and innovation programme under grant agreement No 101073307 (MSCA-DN LEMUR), the European Research Council (ERC) under grant agreement No 951424 (ERC-SyG Water-Futures), the European Union’s Horizon 2020 research and innovation programme under grant agreement No 739551 (Teaming KIOS CoE), and the Republic of Cyprus through the Deputy Ministry of Research, Innovation and Digital Policy.}

The rapid growth of the IoT has significantly advanced Cyber-Physical Systems (CPS), increasing their complexity, and susceptibility to faults. 
In critical infrastructures, like water networks, server machines, and power grids, sensor monitoring is vital for detecting and localizing anomalies to ensure safety and efficiency. However, the growing complexity of CPS makes diagnosing increasingly challenging, requiring advanced algorithms to analyze high-dimensional multivariate time series (MTS) data by capturing both temporal dependencies and inter-correlations across systems. This has driven an increased reliance on Computational/Artificial Intelligence (CI/AI) techniques to manage the complexity of modern CPS and enable intelligent decision-making. Anomaly diagnosis is generally divided into two key tasks: \textbf{anomaly detection}, which identifies whether an abnormal event has occurred, and \textbf{anomaly localization}, which determines the series responsible. Despite the critical importance of anomaly localization, relatively few methods have been specifically developed for addressing this task, compared to the extensive research focused on anomaly detection. The key objective of this paper is to effectively combine the learning of complex dynamics with anomaly localization. Specifically, we contribute to this problem by proposing a methodology for detection and localization of MTS anomalies. \\

Recent advances in Transformer models, driven by their self-attention mechanism \cite{AttentioIsAll}, have demonstrated remarkable effectiveness in processing sequential data, particularly in the domains of natural language processing \cite{devlin-etal-2019-bert}, video and audio analysis \cite{Akbari2021VATTTF}, and computer vision \cite{ViTr}. The inherent ability of Transformers to capture long-range dependencies and model complex interactions makes them especially well-suited for learning the intricate dynamics of MTS \cite{TranMTSReprLearning}. Moreover, Transformers have recently been applied to time series anomaly detection, achieving superior performance and surpassing state-of-the-art methods \cite{TraAD,AnomalyTranformer}. The main contributions of this paper are as follows:

\begin{enumerate}
\item We introduce a \textbf{novel transformer-based localization approach for MTS}. This work is based on a model that captures the complex dynamics of the data, addressing both detection and localization, with high performance (as illustrated in Section \ref{sec:Mehtod}).
    
\item We go beyond black-box models by offering insights into how self-attention mechanisms in transformer-based representations capture temporal and feature dependencies in MTS. In addtion  we connect these findings to statistical models (see Section \ref{sec:Analysis of SALearning}).
    
\end{enumerate}
The paper is organized as follows: Section \ref{sec:Related} reviews the related work, providing a contextual overview. Section \ref{sec:formulation} outlines the problem formulation, followed by Section \ref{sec:Preliminaries}, which covers key preliminaries. Section \ref{sec:Analysis of SALearning} analyzes self-attention learning on MTS, while Section \ref{sec:ProposedMethod} details our proposed method. The experimental setup and results, comparing our approach with state-of-the-art methods, are described in Sections \ref{sec:setup} and \ref{sec:results}, respectively. Finally, Section \ref{sec:Conlusions} concludes the paper by summarizing our findings.
\section{Related work}
\label{sec:Related}
Recent deep learning models have been used with significant success in anomaly detection by leveraging neural networks' ability to capture complex MTS dynamics \cite{su_robust_2019,li_multivariate_2021} \cite{AnomalyTranformer}. Despite these advances, anomaly localization has remained relatively underexplored but has recently begun to attract more investigation, as detection methods alone fail to capture the complex dynamics of large-scale systems.\\

In this context, the MSCRED \cite{MSCRED} model constructs correlation matrices from MTS data, processed through a convolutional LSTM-based encoder-decoder, and localizes anomalies based on residual magnitudes. Similarly, Liang et al.\cite{9666863} enhance this approach with adversarial training, while Choi et al.\cite{Choi2020GANBasedAD} use 2D matrices based on time series distances.
\cite{DAEMON,DAEMON2} introduced an adversarial autoencoder with dual discriminators that detects anomalies based on total reconstruction error and localizes them by integrating the gradients for each dimension of an anomalous entity \cite{IGradients}. A more advanced representation learning model, called OmniAnomaly \cite{su_robust_2019}, that aims to address the localization task, employs a gated recurrent unit (GRU)-based autoencoder combined with a variational autoencoder (VAE) to detect anomalies and interpret them by ranking the contribution of each time series dimension to the anomaly score. Building on this progress, Li et al. \cite{li_multivariate_2021} proposed InterFusion, a hierarchical VAE that models inter- and intra-dependencies in MTS. It uses Markov chain Monte Carlo  (MCMC) to refine the reconstruction error at anomalous time steps, which is then used as the anomaly score for each time series. However, due to computational complexity, it is limited to interpreting anomaly segments rather than individual time steps.\\
 
Some researchers have used simpler, more interpretable methods for anomaly localization, improving explainability but reducing detection accuracy, which limits their effectiveness and real-world applicability. For instance, ARCANA \cite{ARCANA} employs an optimization-based solution with a basic autoencoder, but its difficulty to capture the complex dynamics of MTS limits its overall performance in some practical applications. 
 Another technique designed to offer a more `clear' anomaly score is counterfactual reasoning \cite{CounterfactualReasoning}, where variables in anomalous intervals are replaced with normal data to identify key contributing variables based on whether the anomaly persists. However, this substitution-based approach may not be reliable to reproduce possible non-stationarities. Additionally, Explainable AI (XAI) methods like SHapley Additive exPlanations (SHAP) have been explored \cite{10069107,9568906}. Despite their utility, SHAP calculations are computationally expensive and rely on assumptions like feature independence, which may not hold in MTS. Approximation methods like Kernel SHAP reduce costs but still rely on sampling and data substitution, which generally do not hold for MTS. \\

While Transformers are widely used for detection and forecasting \cite{AnomalyTranformer,1Autoformer}, their potential for anomaly localization is underexplored. Recently, they have also started gaining attention in research for diagnostic tasks in MTS. For example, Wang et al. \cite{10405930} used attention mechanisms to detect anomaly start points, while Wu et al. \cite{WU2023439} applied Transformers for fault classification in rotary systems. However, precise anomaly localization remains underexplored. Deep learning models perform well at capturing complex dynamics for detection but often lack interpretability, thus complicating localization. However, the Transformer encoding process can be effectively integrated with statistical models, thereby enhancing both detection and localization, as discussed in Section \ref{sec:Analysis of SALearning}

\section{Problem formulation}
\label{sec:formulation}
\subsection{Anomaly diagnosis in MTS}
A discrete univariate time series $x \in \mathbb{R}^{T}$ is a sequence of observations \( x_t \) recorded at discrete times \( t \in T_o \), with $T_o$ denoting the set of observation times. When multiple univariate time series are observed jointly, they form a MTS, denoted as \( X = \{ x^1, x^2, \dots, x^d \} \), where each column vector \( x^i \in \mathbb{R}^T \) represents the \( i'th \) time series, the row vector $x_t \in \mathbb{R}^d $ is the MTS at time step $t$, and \( x_t^i \in \mathbb{R} \) is the value at time \( t \) of the \( i'th \) series. Anomaly diagnosis involves two primary tasks: (i) identifying an unseen observation $x_t$, as anomalous if it deviates in some sense, from historical normal patterns; and (ii) determining which time series are anomalous. This can be formulated mathematically as follows: \\

\begin{definition}[\textbf{Anomaly Detection}]
 Define a function \(\mathcal{F}_D: \mathbb{R}^d \rightarrow \{0,1\}\) that takes as input a MTS \(x_t\) at time step \(t\), and performs binary classification, assigning it either to the normal class (\(y_t = 0\)) or the anomalous class (\(y_t = 1\)). \\
\end{definition}

For the localization task, we formulate the process as a three-stage approach, mirroring decision-making processes in real-world applications: time-step-wise, window-based, and segment-based localization.\\ 

\begin{definition}[\textbf{Time-step-wise Localization}]
Define a function $(\mathcal{F}_L: \mathbb{R}^d \rightarrow [0,1]^d$ that takes as input a MTS \(x_t\) at time step \(t\) and outputs an anomaly score vector \(\mathcal{F}_L(x_t) = (AS_t^1, \dots, AS_t^d)\), where each $AS_t^i$ represents the anomaly score for the \(i'th\) time series at time step \(t\), taking values from 0 to 1.\\
\end{definition}
These scores quantify the probability of each time series being anomalous, and anomalous series can be identified by applying a threshold. This method evaluates each time step independently, without delay or the need to store past values. However, relying solely on time-step-wise localization can be insufficient for complex systems. To address this, we extend anomaly localization to encompass broader contexts.\\

\begin{definition}[\textbf{Window-Based Localization}]  
The window-based anomaly score (WAS) for the \(i'th\) time series at time step \(t\) is computed by aggregating anomaly scores over a window \(W_t\) around \(t\), where \(W_t = [t - w_1, t + w_2]\). Formally, it is defined as: $\mathrm{WAS}_t^i = \text{agg}_{\tau \in W_t} AS_\tau^i$. A reasonable choice for the \(\text{agg()}\) is the $max()$ function.\\
\end{definition}
The parameters $w_1$ and $w_2$ are the \textit{look-back} and \textit{look-ahead} window sizes, respectively, defining how past and future data are considered in the localization process. When \(w_2 = 0\), decisions are made in real-time with minimal memory use. By defining \(W_t\) to include all consecutive anomalous time steps, we derive \textbf{Segment-based Localization}. 


\section{Preliminaries}
\label{sec:Preliminaries}
The attention mechanism is a key neural network component that captures dependencies between input tokens by measuring their similarities. Empirical evidence suggests it is more effective when applied to data embedded in higher-dimensional space, akin to how Support Vector Machines (SVMs) \cite{SVM} use higher-dimensional embeddings to linearize complex relationships. In multivariate time series analysis, tokens represent the MTS at specific time points \( x_t \in \mathbb{R}^d \), for \( t = 1,2 \dots, T \). Each \( x_t \) is embedded into a vector of dimension \( d_{\text{model}} \), and a positional encoding vector \( \bm{P} \in \mathbb{R}^{d_{\text{model}}} \) (either learnable or deterministic) is added to each token to incorporate temporal information. This results in a dataset \( \bm{X}^{(0)} \in \mathbb{R}^{T \times d_{\text{model}}} \), which serves as the input for subsequent attention-based computations. At each layer \( l \), the self-attention mechanism operates by projecting the input \( \bm{X}^{l-1} \in \mathbb{R}^{T \times d_{\text{model}}} \) into the Query($\mathcal{Q}^l$), Key($\mathcal{K}^l$), and Value ($\mathcal{V}^l$) matrices using the corresponding learnable projection matrices \( W_{\mathcal{Q}}^l = \{w_{ij}^Q\} \), \( W_{\mathcal{K}}^l = \{w_{ij}^K\} \), and \( W_{\mathcal{V}}^l = \{w_{ij}^V\} \), all of which are of dimension \( \mathbb{R}^{d_{\text{model}} \times d_{\text{model}}} \). These projections are computed as:
\begin{equation}   
\mathcal{Q}^l = \bm{X}^{l-1} W_{\mathcal{Q}}^l, \quad \mathcal{K}^l = \bm{X}^{l-1} W_{\mathcal{K}}^l, \quad \mathcal{V}^l = \bm{X}^{l-1} W_{\mathcal{V}}^l. 
\end{equation}

Here, \( \mathcal{Q} = \{q_t^j\} \), \( \mathcal{K} = \{k_t^j\} \), and \( \mathcal{V} = \{v_t^j\} \in \mathbb{R}^{T \times d_{\text{model}}} \). These are used in the self-attention mechanism to derive the attention scores of the $l'th$ layer, \( S^l = \{S^l_{ij}\} \), which are computed as:

\begin{equation} \label{eq:Self-Attention Weights}
S^l = \text{softmax}\left( \frac{\mathcal{Q}^l (\mathcal{K}^l)^{T}}{\sqrt{d_{\text{model}}}} + M \right) \in \mathbb{R}^{T \times T} ,
\end{equation}
where \( M \) is an optional mask, and the scaling factor \( \sqrt{d_{\text{model}}} \) ensures numerical stability by normalizing the dot product between the query and key matrices \cite{AttentioIsAll}. The matrix \( M \in \mathbb{R}^{T \times T} \) is used to implement masking, ensuring that the latent representation can only attend to preceding time steps, thereby maintaining its causality property. The final step is the multiplication of the self-attention matrix \( S^l \) with the value matrix \( \mathcal{V}^l \), resulting in the latent representation
\begin{equation}\label{eq:LatentSA}
    Z^l = S^l \mathcal{V}^l = \text{Attention}(\mathcal{Q}^l, \mathcal{K}^l,\mathcal{V}^l) \in \mathbb{R}^{T \times d_{model}} . 
\end{equation}
In practice, multi-head attention (MHA) is used to capture complex patterns by performing multiple parallel attention operations (heads), each with its own projection matrices \( W_{\mathcal{Q}}^l \), \( W_{\mathcal{K}}^l \), and \( W_{\mathcal{V}}^l \), sized \( \mathbb{R}^{d_{\text{model}} \times \frac{d_{\text{model}}}{H}} \), where $H$ is the total number of heads. For each head, the latent representation \( Z_h, h=1\cdots,H \), is computed using Equation \eqref{eq:LatentSA}. These are concatenated and once again projected using \( W^O \in \mathbb{R}^{d_{model} \times d_{model}} \). The final latent representation is computed as:
\begin{equation}
Z = \text{MHA}(Z_1, Z_2, \dots, Z_H) = \text{Concat}(Z_1, \dots, Z_H) W^O.
\end{equation}

\section{Analysis of representation learning of attention mechanism on MTS}
\label{sec:Analysis of SALearning}

\subsection{Unveiling hidden representation of MTS}

In the context of MTS analysis, understanding the roles of  \(Q\), \(K\), and \(V\) is essential for effective anomaly diagnosis. Each column of \(Q\), \(K\) is determined by the following equations
\begin{subequations}
\begin{equation}\label{eq:qj}
    q_t^{j} = w_{jj}^Q x_t^{j} + \sum_{m \neq j} w_{mj}^Q x_t^{m},
\end{equation}
\begin{equation}\label{eq:kj}
    k_t^{j} = w_{jj}^K x_t^{j} + \sum_{m \neq j} w_{mj}^K x_t^{m},
\end{equation}
\end{subequations}
where \(t = 1, \ldots, T\) and \(j = 1, \ldots, d\). The weights \(w_{mj}^Q\) and \(w_{mj}^K\) are derived from the learned projection matrices \(W_Q\) and \(W_K\), respectively. Each time series is thus transformed into a \textbf{linear combination of itself and the other time series at the same time step}. Equations (\ref{eq:qj}) and (\ref{eq:kj}) can be viewed as parallel to two statistical models. Firstly, they bear resemblance to a linear regression model. Secondly, they align with the Spatial Autoregressive (SAR) model, where different time series are treated as spatial dimensions \cite{cressie2011statistics}.
Using Equation \eqref{eq:Self-Attention Weights}, the (masked) self-attention weights are calculated, determining the attention that the MTS at time step \( t \) should allocate to previous time steps.\\

The final latent representation, computed by \ref{eq:LatentSA} is given by:
\begin{equation}\label{eq:LRep}
z^j_t= w^V_{jj}\left( \sum_{k=1}^t s_{tk}x^j_k \right) +\sum_{m \neq j} \left[ w^V_{mj}\left( \sum_{k=1}^t s_{tk}x^m_k \right)\right].
\end{equation}
From the expression (\ref{eq:LRep}), we observe that each time series in the latent space is represented as a double-weighted average. The first component corresponds to an autoregressive process \((\text{AR}(t))\) of the time series itself, while the second component involves \(d_{model}-1\) autoregressive processes \((\text{AR}(t))\) of the other time series, which can be interpreted as explanatory variables contributing to the behavior of the \(j'th\) time series. The influence of each time series is determined by the weights \(w_{ml}\), where \(m, l=1,\ldots,d_{model}\), which are derived from the projection matrix \(V\). The latent representation in \eqref{eq:LRep} aligns with the Space-Time Autoregressive (STAR) model \cite{cressie2011statistics}, capturing temporal dependencies and treating time series as interacting spatial dimensions.

\subsection{Effect of anomaly on the latent representation} \label{sub:Effect}
\begin{figure}[ht]
    \centering
    \includegraphics[width=1\linewidth,height=5.4cm]{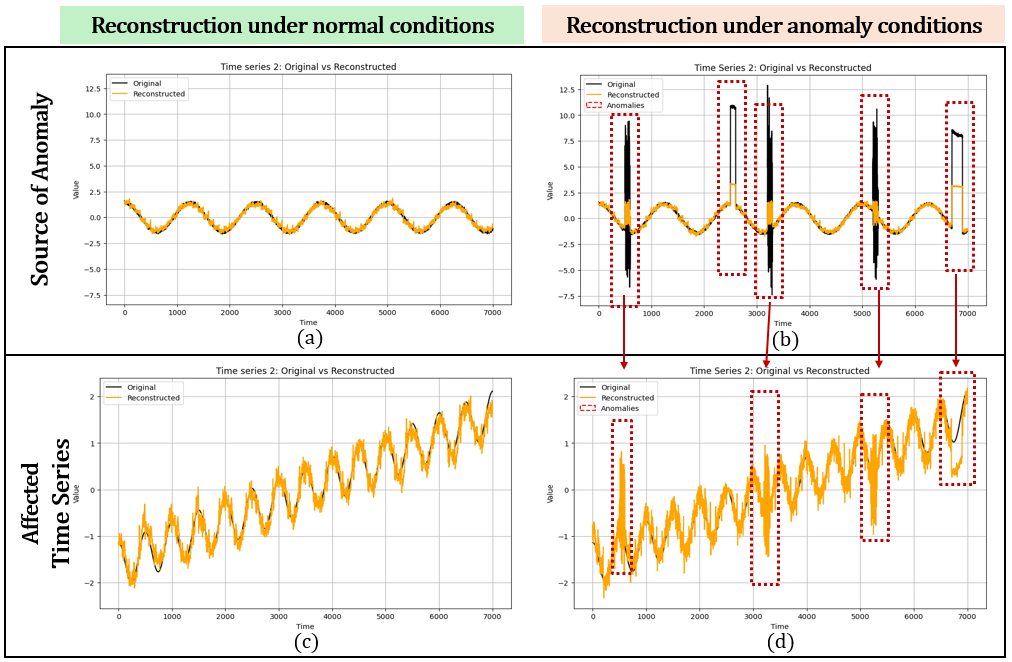}
    \caption{Subplots (a) and (c) show the reconstructions of Time Series 1 and Time Series 2 under normal conditions, with the two series being uncorrelated (Spearman, Pearson, and Kendall's Tau correlations all equal to zero). On the right, subplot (b) highlights anomalies in Time Series 1 (red boxes), while subplot (d) illustrates how these anomalies affect the reconstruction of the normal Time Series 2.}

    \label{fig:Spread}
\end{figure}

To illustrate the effect of an anomaly on the latent representation of the entire MTS, we assume that only one time series, \( x^a \), is anomalous. The anomalous time series is modeled as:
\begin{equation}
    x^a = 
\begin{cases}
x_t^{normal} + A(t), & \text{if } t \in [t_1, t_2], \\
x_t^{normal}, & \text{otherwise},
\end{cases}
\end{equation}
where \( A(t) \) is an additive anomaly function active during the interval \([t_1, t_2]\), and \( x_t^{normal} \) represents the normal component of the time series. Then the latent representation of the $j'th$ time series at each time step, \( z^j_t \), is given by:
\begin{equation}
    z^j_t = \sum_{k=1}^{t} s_{tk} \text{v}^j_k + 
    \underbrace{\sum_{k=t_1}^{t} w^{V}_{aj} A(k)}_{\text{Anomaly effect}}.
    \label{eq:Spread}
\end{equation}
Equation \eqref{eq:Spread} reveals that anomalies in one time series can have an impact to the latent representations of the other time series in the dataset, \textbf{with the extent of this influence being governed by the weight \( w^{V}_{aj} \)}. Ideally, if these weights reflected actual correlations, anomalies would only impact correlated series. However, as confirmed by our experiments and shown in Figure \ref{fig:Spread}, empirical observations indicate this often doesn't hold. Specifically, in Figure \ref{fig:Spread}, we show how an anomaly in one time series affects the reconstruction of an uncorrelated series, using the transformer encoding-decoding process (as detailed in Section \ref{sub:RLM}). 

\section{Proposed method}
\label{sec:ProposedMethod}
\begin{figure}[ht]
    \centering
    \includegraphics[width=1\linewidth,height=4cm]{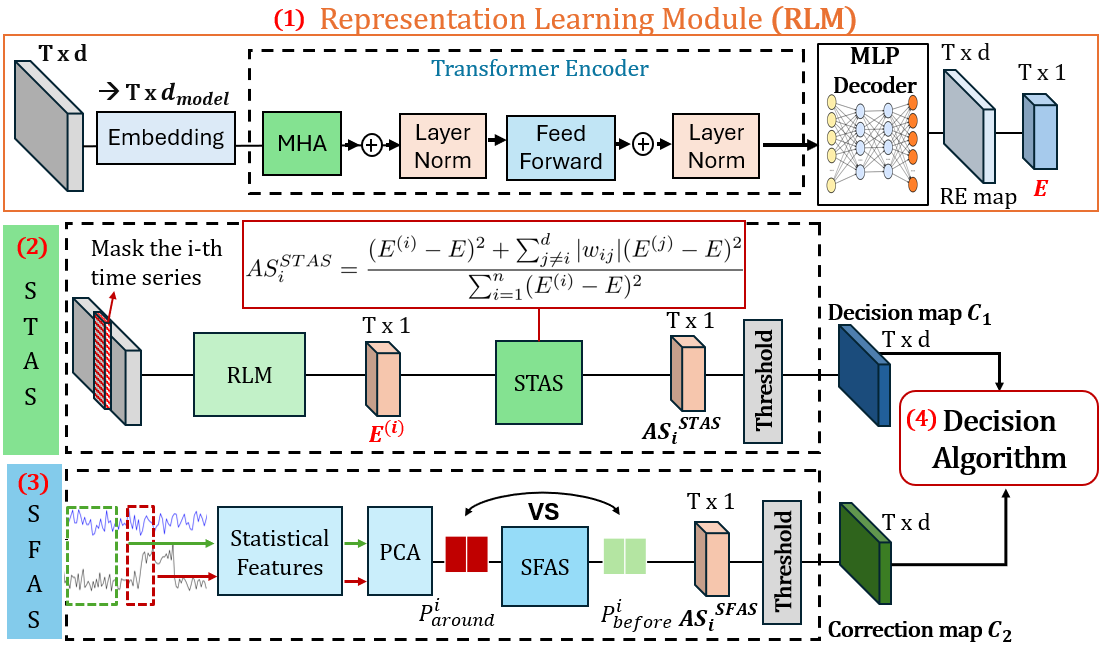}
    \caption{Overview of the proposed MTS anomaly localization method}
    \label{fig:overview}
\end{figure}

The proposed model comprises four key components, as illustrated in Figure \ref{fig:overview}. The first component, the Representation Learning Module (RLM), employs a transformer-based encoder to capture the complex dynamics of MTS, with the encoded representations processed by a multi-layer perceptron (MLP) for reconstruction (see Section \ref{sub:RLM}). The next two components are the STAS (2) and SFAS (3) modules, which are proposed to address the anomaly localization task. STAS utilizes the RLM (Section \ref{sub:stas}), while SFAS focuses on the most significant changes in key statistical features (Section \ref{sub:sfas}). Lastly, the outputs of STAS and SFAS are integrated through the proposed decision Algorithm \ref{alg:stas_localization}, to yield the final anomaly localization decision.

\label{sec:Mehtod}
\subsection{Representation learning}
\label{sub:RLM}
Considering the streaming characteristics of time series data, the recorded MTS is processed in non-overlapping windows of length $T$. After embedding the input MTS into $X^0 \in \mathbb{R}^{T \times d_{\text{model}}}$, the transformer encoder, consisting of $L$ layers, captures the underlying dynamics of the MTS through the following process. At each layer $l$, the output is computed as:

\begin{equation}
\begin{aligned}
    Z^l &= \text{Layer-Norm}(\text{MHA}(X^{l-1}) + X^{l-1}) \\
    X^l &= \text{Layer-Norm}(\text{Feed-Forward}(Z^l) + Z^l).
\end{aligned}
\end{equation}
First, multi-head attention (MHA) is applied to the output of the  previous layer \( X^{l-1} \), followed by a residual connection and layer normalization, producing \( Z^l \). Then, \( Z^l \) passes through a feedforward neural network, with another residual connection and normalization, resulting in the final representation \( X^l \) for layer \( l \). The final output, denoted as \( X^L \), from the last layer \( L \) is passed through a multi-layer perceptron (MLP), which produces the reconstructed MTS, represented as \( \widehat{X} \in \mathbb{R}^{T \times d} \).
The objective function \( \mathcal{L}_{\text{Total}} \) used to train the model includes two key components:
\begin{equation}
\mathcal{L}_{\text{Total}}(\widehat{X}, P, S, \lambda; X) 
= \|X - \widehat{X}\|_{\text{F}}^2 - \lambda \|D_{\text{div}}(P, S)\|_{1}.
\end{equation}
The first term minimizes the Frobenius norm of the reconstruction error, ensuring accurate reconstruction of the time series. The second term represents the $L_1$ norm of the distribution divergence (\( D_{\text{div}} \)), introduced by \cite{AnomalyTranformer} with \( \lambda \) as the trade-off parameter balancing the loss terms. The quantity $D_{\text{div}} \in \mathbb{R}^{T \times 1}$ is computed as follows: For each layer, except for learning the self-attention matrix \( S^l \in \mathbb{R}^{T \times T} \), the models also learn the prior-attention matrix \( P^l \in \mathbb{R}^{T \times T} \). In both matrices, each row \( S^l_t \in \mathbb{R}^{1 \times T} \) and \( P^l_t \in \mathbb{R}^{1 \times T} \) represents the discrete distribution of the weights that time step \( t \) assigns to previous time steps. The prior attention distribution for each time step $t,$ \( P^l_t \), is modelled using a Laplace kernel, following the method from Bai et al. \cite{bai_paformer_2023}. Then Kullback-Leibler (KL) divergence measures, the difference between $S^l_t$ and the $P^l _t$ distribution for each time step creating the $T$-dimensional vector $D_{div}$.
\begin{equation}
    D_{\text{div}}(P,S) = \frac{1}{L} \sum_{l=1}^{L} \left( \mathrm{KL}( P^l_t \| S^l_t) + \mathrm{KL}(S^l_t \| P^l_t) \right).
\end{equation}
This encourages the self-attention mechanism to focus less on adjacent time points, which reduces the risk of overfitting to potential anomalies that may be present in the training set. The parameters of the Laplace kernels $P^l_t$ (for $t=1,\dots,T$) are learnable, and at each layer are initialized by setting them equal to the self-attention distributions from the previous layer, $S^{l-1}_t$. The anomaly detection score at each time step is computed as the elementwise multiplication (\(\odot\)) between the \(L_2\) norm (\(\| \cdot \|_2\)) of the total reconstruction error, \( E_t = \| x_t - \hat{x}_t \|_2^2 \), and a score derived from the divergence measure \( D_{div} \), following the approach used in \cite{bai_paformer_2023,AnomalyTranformer}.
\begin{equation}\label{eq:AnomalyscoreDetection}
\mathrm{AS}(x_t) = \|x_t - \hat{x_t}\|_2^2 \odot  \mathrm{Softmax}\left(-D_{div}(P, S)\right) \in \mathbb{R}^{T \times 1}.
\end{equation}

The computed anomaly score is then passed through a one-sided FIR-CUSUM (Fast Initial Response Cumulative Sum) algorithm \cite{FIRCusum} to detect anomalies:
\begin{align}
    CS_t &= \max \left(0, AS_t - (\mu + K) + CS_{t-1}\right), \quad CS_0 = b.
\end{align}

where \( \mu \) is the expected value of anomaly scores under normal conditions, typically set to 0. An anomaly is detected when the cumulative sum \( CS_t \) exceeds a decision threshold \( n\sigma \), where \( n \) is a dataset-specific parameter and \(\sigma\) is the standard deviation of \( CS_t \) computed over the training set.

\subsection{STAS}
\label{sub:stas}

The analysis in Section \ref{sub:Effect} reveals two key findings: (i) anomalies in one time series can affect uncorrelated series, obscuring the true source of the anomaly, and (ii) the learned latent representations in the self-attention process are closely connected to the Space-Time Autoregressive (STAR) model. The STAS module is designed to generate a localization anomaly score that mitigates the anomaly effect and leverages the connection to space-time models. We apply the representation learning module to the MTS, masking the \( i'th \) series, which produces \( E^{(i)}_t \), the total reconstruction error at time step \( t \) without the influence of the \( i'th \) series. The anomaly score for the \( i'th \) series, at the (anomalous) time step $t$, is computed as the sum of two components: (i) the individual contribution, given by \( (E^{(i)}_t - E_t)^2 \), representing the change in total reconstruction error when the \( i'th \) series is masked; and (ii) the contribution from correlated series, calculated as the weighted sum of squared deviations \( (E^{(j)}_t - E_t)^2 \) from other series, weighted by their absolute correlation \( |w_{ij}| \) with the \( i'th \) series. The final score is normalized by the total squared deviations across all series. The vector $AS_i^{STAS} \in \mathbb{R}^{{T}}$ contain the anomaly scores of the $i'th$ time series at all (anomalous) time step and is defined as:

\begin{equation}\label{eq:Space Time Lisa metric}
AS^{STAS}_i = \frac{(E^{(i)} - E)^2 + \sum_{j \neq i}^{d} |w_{ij}| (E^{(j)} - E)^2}{\sum_{i=1}^{n} (E^{(i)} - E)^2},  \quad i=1,\cdots d
\end{equation}
We define this metric, inspired by the Space-Time Local Indicators of Spatial Association (LISA), commonly used to detect outliers in space-time models such as STAR \cite{TAO2023102042}. This definition allows higher anomaly scores in dimensions correlated with the true source of anomalies. Additionally, it ensures that false alarms are more likely to occur in time series that, while possibly normal, share a dependency with an anomalous series. The weights \( w_{ij} \) signify the correlation between the \( i'th \) and \( j'th \) time series. We primarily use the Spearman correlation matrix, as it demonstrated better performance across all datasets. Similar results were achieved with Kendall’s Tau. Pearson correlation was not applied due to its limitation in revealing only linear correlations. The localization decision map from the STAS metric, named \( C_1 \in \mathbb{R}^{T\times d}\), is computed by marking a time series as anomalous if its \( AS^{STAS}_i \) exceeds a predefined threshold ($\Tilde{h}_1$).
\subsection{SFAS}
\label{sub:sfas}
To further enhance localization performance and address complex anomalies that cause false positives and negatives, we also introduce the Statistical Feature Anomaly Score (SFAS) as a corrective factor to the STAS. SFAS quantifies abnormality by analyzing statistical features of time series within windows before (\(W_{\text{before}}\)) and around (\(W_{\text{around}}\)) detected anomalies. Specifically, we extract \(k\) statistical features, resulting in matrices \(F_{\text{before}} \in \mathbb{R}^{k \times d}\) and \(F_{\text{around}} \in \mathbb{R}^{k \times d}\), where \(d\) represents the number of time series. For \textit{time step-wise localization}, features are extracted up to time \(t\), while for \textit{window-based localization}, they are extracted from the window surrounding \(t\). Key statistical features include variance, trend strength, linearity, curvature, seasonality, and other important data characteristics \cite{LargeScaleOnlineAS}. After applying Principal Component Analysis (PCA) for dimensionality reduction, we obtain two-dimensional projections \(P_{\text{before}}\) and \(P_{\text{around}}\) in \(\mathbb{R}^{2 \times d}\). The SFAS anomaly score for the \(i'th\) time series is calculated as the \(L_1\)-norm of the difference between \(P_{\text{before}}^i\) and \(P_{\text{around}}^i, \text{both in }\mathbb{R}^{2}\).
\begin{equation}
\label{eq:SFAS}
    AS^{SFAS}_i = \| P_{\text{before}}^i - P_{\text{around}}^i \|_1 \in \mathbb{R}^T.
\end{equation}
As detailed in Algorithm~\ref{alg:stas_localization}, the Space-Time Anomaly Score (STAS) is initially employed for anomaly classification producing the decision map $C_1$. If the SFAS for a given time series surpasses a predefined threshold ($\Tilde{h}_2$) after being previously classified as normal, SFAS is applied for reclassification. This iterative process allows for more accurate decision-making, minimizing potential misclassifications.

\begin{algorithm}
\small
\caption{STAS/SFAS Localization}
\label{alg:stas_localization}

\textbf{Input:}

Transformer model parameters: $d_{\text{model}}$, $L$, $H$ 

Data: $X \in \mathbb{R}^{T \times d_{\text{model}}}$ : MTS 

From detection: $E \in \mathbb{R}^{T}$: Total reconstruction error

\SetKwFunction{FLocalization}{\textbf{STAS/SFAS}}
\SetKwProg{Fn}{Function}{:}{}
\Fn{\FLocalization{$E$,X}}{

    \For{each time series $TS_i$}{
         $X^{(i-masked)}$: The MTS with $i'th$ time series masked
    RLM($X^{(i-masked)}$), get $E^{(i)}$}
    
        Compute $AS^{STAS}$ (Eq.~\eqref{eq:Space Time Lisa metric});
        
        \textbf{If} $AS^{STAS}_i > \Tilde{h}_1$ $\mathbf{C}_{1,i} \gets 1$ \textbf{Else:} $\mathbf{C}_{1,i} \gets 0$

    $p \gets 0$ (Number of corrections), $\mathbf{C}_{\text{combined}}=\mathbf{C}_{\text{1}}$
    
    \For{each time series $i$}{
$\mathbf{P}_\text{before}\!=\!\text{PCA}(F_\text{before})$, $\mathbf{P}_\text{around}\!=\!\text{PCA}(F_\text{around})$

Compute $AS^{SFAS}$ using Eq.~\eqref{eq:SFAS} 

        \textbf{If} $AS^{SFAS}_i > \Tilde{h}_2$ \textbf{and} $\mathbf{C}_{\text{1},i} = 0$ 
        
            $\mathbf{C}_{\text{combined,i}},\mathbf{C}_{\text{2},i} \gets 1$; $p \gets p+1$ \textbf{Else:} 
            $\mathbf{C}_{\text{2},i} \gets 0$
    }

    sorted\_anomalous $\gets$ sort indices of $\mathbf{C}_{\text{1},i} = 1$ by ascending $AS^{STAS}_i$;
    
\For{$j = 1$ to $p$}{ 
    $\mathbf{C}_{\text{combined},\text{sorted\_anomalous}[j]} \gets 0$ 
}

    \KwRet{$\mathbf{C}_{\text{1}}, \mathbf{C}_{\text{2}}, \mathbf{C}_{\text{combined}} \in \mathbb{R}^{T\times d}$, the localization decision maps}
    }
\end{algorithm}

\section{Experimental setup}
\label{sec:setup}

The transformer module uses a model dimension of $d_{\text{model}} = 512$, 8 attention heads, 3 layers, non-overlapping windows of $T = 100$, and λ=3. The ADAM optimizer is applied with a learning rate of $10^{-4}$, with early stopping to prevent overfitting. The threshold $\Tilde{h}_1$ is set at the $(1-a_t/d)\%$ percentile of $AS^{STAS}$, where $a_t$ is the ground truth number of anomalous dimensions. For each segment, $\Tilde{h}_2$ is set as the 93rd to 99th percentile of $AS^{SFAS}$, depending on the dataset, using data up to time $t$. We report mean performance metrics and standard deviations from multiple experiments for all results.

\subsection{Comparisons}
We compare our approach with the following state-of-the-art (SOTA) methods:
\begin{itemize}
    \item \textbf{DAEMON} (DA)\cite{DAEMON}: Adversarial Autoencoder  where the  integrating gradients (IG) \cite{IGradients} method is used for localization.
    \item \textbf{OmniAnomaly} (OA) \cite{su_robust_2019}: A GRU-based VAE autoencoder where the reconstruction error is utilized for both anomaly detection and localization.
    \item \textbf{InterFusion} (IF) \cite{li_multivariate_2021}: A hierarchical GRU-VAE model that enhances localization by applying MCMC techniques to refine reconstruction errors at anomalous time steps.
\end{itemize}
The IF model is limited to segment-based localization due to the high computational cost of MCMC, allowing for comparison only with this localization type.

\subsection{Datasets}
\textbf{Synthetic Dataset:} 
\begin{itemize}
    \item \textbf{Waves (WVS)}: The WVS dataset consists of 10 time series, grouped by shared frequency. Each series is generated using the equation: $
    X_t = B\sin(2\pi \omega t + \phi) + \epsilon_t$, where \( B \) is the amplitude, \( \omega \) is the frequency, \( \phi \) is the phase shift, and \( \epsilon_t \sim N(0, 1) \). The four frequencies \( \omega = 10^{-5}, 10^{-4}, 10^{-3}, 10^{-2} \) are assigned to groups: one series in the first group, and three in each of the remaining groups. \( B \) is randomly selected from [2, 3], and \( \phi \) from \( [0, \frac{\pi}{2}] \). Anomalies are introduced at intervals, either by adding sine waves with deviating frequencies or injecting constant-value outliers.
\end{itemize}
\textbf{Real-World Datasets:}

\begin{itemize}
  \item \textbf{Server Machine Dataset (SMD)} \cite{su_robust_2019}: This dataset contains 38 time series of server metrics, including memory usage, CPU utilization, and ETH inflow, collected over 5 weeks from a large internet company.
    \item \textbf{Application Server Dataset (ASD)} \cite{li_multivariate_2021}: The ASD dataset is composed of MTS data collected from a large internet company. It includes 19 metrics recorded every 5 minutes, primarily from virtual machines and servers.
\end{itemize}
All datasets are imbalanced in terms of normal and anomalous time steps, requiring the use of appropriate evaluation metrics.

\subsection{Evaluation Metrics}
We use standard evaluation metrics, including Precision$=\left( TP/(TP + FP) \right)$, Recall$=\left( TP/(TP + FN) \right)$, and F1-Score=\( \left( \frac{2 \times \text{Precision} \times \text{Recall}}{\text{Precision} + \text{Recall}} \right) \).
 Here, $TP$ refers to correctly identified anomalies, $FP$ to false positive identifications, and $FN$ to missed anomalies. We also compute the Area Under the Curve (AUC), which measures the performance of the models by plotting the true positive rate against the false positive rate across various thresholds. For evaluating the segment-based localization we also utilize the Interpretation Score (IPS) \cite{li_multivariate_2021}, which assesses how accurately the model localizes anomalous segments. It computes the weighted proportion of correctly predicted anomalous dimensions across segments as  $ IPS = \sum_{i=1}^{N} \left( {W_i |G_{S_i} \cap P_{S_i}|}\right)/\left({|G_{S_i}|}\right)$, where $N$ is the total number of anomalous segments, $W_i$ is the weight for each segment (equally weighted). $G_{S_i}$ and $P_{S_i}$ represent the ground truth and predicted anomalous dimensions for segment $S_i$, respectively.


\section{Experimental results}
\label{sec:results}
\subsection{Empirical Analysis}  
\begin{figure}
    \centering    \includegraphics[width=1\linewidth,height=3.5cm]{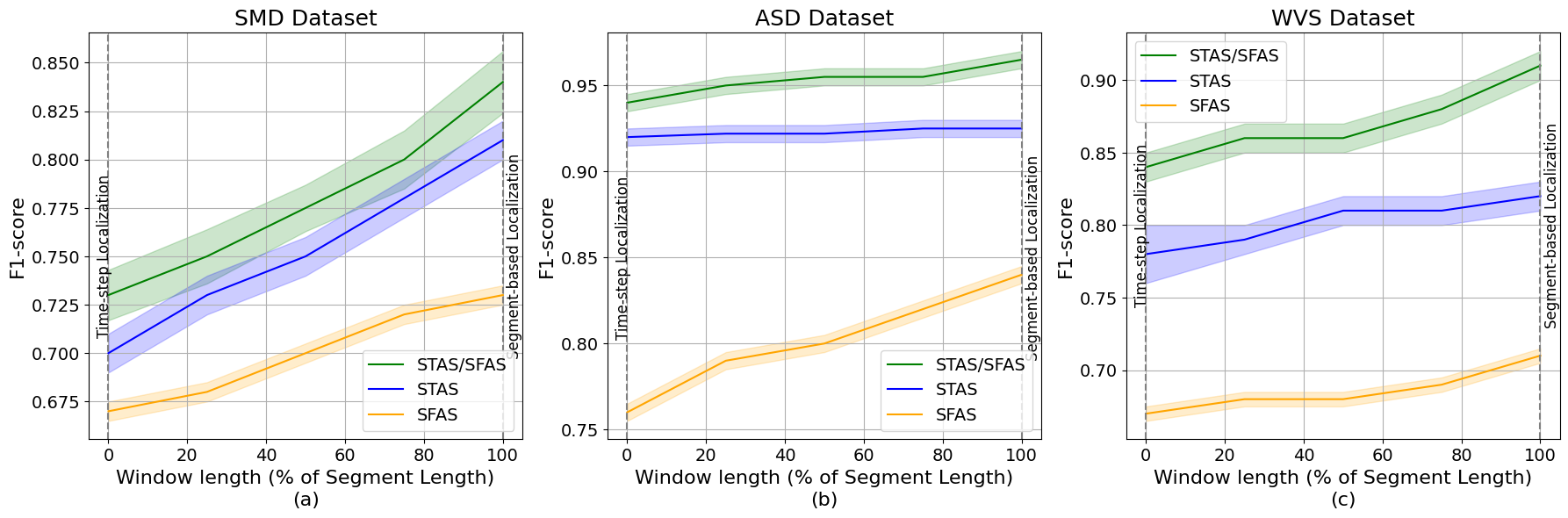}  
    \caption{Comparison of F1-scores using STAS, SFAS, and their combination for anomaly localization across the three datasets.}
    \label{fig:comparison_f1_scores_combined}
\end{figure}

In this analysis, we examine how incorporating SFAS as a correction term improves anomaly localization. Figure \ref{fig:comparison_f1_scores_combined} demonstrates the improvement in anomaly localization across different window lengths when using SFAS as a correction term for STAS within the proposed Algorithm \ref{alg:stas_localization}. As the window length increases in window-based localization, the SFAS correction term becomes more effective, as larger windows enable more accurate capture of statistical features. 

\subsection{Results}
In this section, we evaluate the performance of our proposed localization methodology, comparing it to recent SOTA methods using the selected datasets. For the real-world datasets, all methods demonstrated high detection performance (F-1 score $> 95\%$), although OmniAnomaly and Daemon displayed lower reconstruction quality. In the synthetic dataset Daemon showed lower performance and is therefore excluded from comparison. The presentation of localization results is structured to reflect the decision-making process in real-world CPS, where information is progressively revealed over time.\\ 

\begin{figure}
    \centering
   \includegraphics[width=1\linewidth,height=3.5cm]{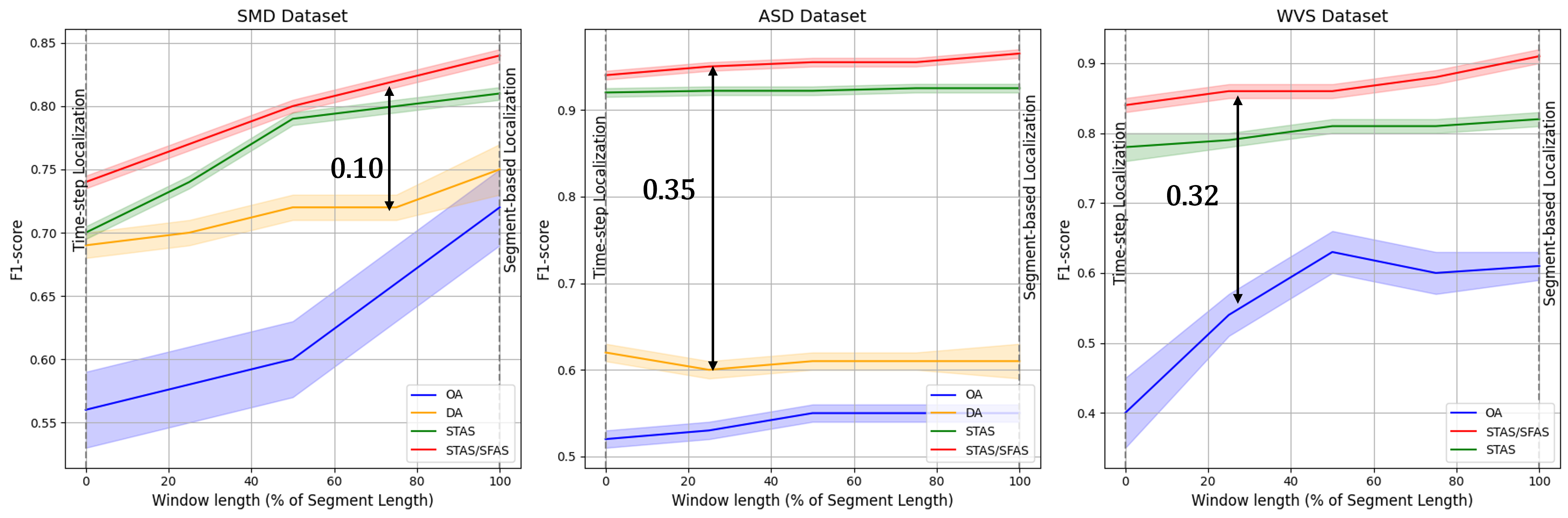}
   \caption{F1-scores of all methods across different window lengths for window-based localization, with arrows showing the maximum improvement over the best-performing method.}
    \label{fig:SegmentLoc}
\end{figure}

First, we focus on time-step-wise localization, which is crucial for near-real-time decision-making. As shown in Table~\ref{tab:TimeLoc}, our model achieves a 9\% improvement in time-step-wise localization on the SMD dataset. On the ASD dataset, despite all methods achieving similar detection performance, our model demonstrates a 55\% improvement in localization, highlighting its ability to excel in both detection and localization simultaneously. 
\begin{table}
    \caption{Evaluation of Time Step-wise Localization}
        \label{tab:TimeLoc}
    \hspace{-0.25cm}
\scalebox{1.0}{ 
    \centering
    \setlength{\tabcolsep}{3pt}  
    \renewcommand{\arraystretch}{1.1}  
    \begin{tabular}{>{\raggedright\arraybackslash}p{0.02\textwidth}|c|c|c|c|c}  
    \toprule
         \multicolumn{1}{l|}{\textbf{Dataset}} & \textbf{Model} & \textbf{F-1} & \textbf{P} & \textbf{R} & \textbf{AUC} \\
    \midrule
    \multirow{4}{*}{\rotatebox[origin=c]{90}{ASD}}
     & DA  & 0.62\textpm{0.1} & 0.62\textpm{0.1} & 0.62\textpm{0.1} & 0.45\textpm{0.1} \\
       & OA  & 0.52\textpm{0.1} & 0.52\textpm{0.1} & 0.52\textpm{0.1} & 0.29\textpm{0.1} \\
       & STAS    & \textbf{0.92\textpm{0.01}} & \textbf{0.92\textpm{0.01}} & \textbf{0.92\textpm{0.01}} & \textbf{0.89\textpm{0.01}}  \\
       & STAS/SFAS    & \textbf{0.96\textpm{0.01}} & \textbf{0.96\textpm{0.01}} & \textbf{0.96\textpm{0.01}} & \textbf{0.94\textpm{0.01}}  \\
    \midrule
    \multirow{4}{*}{\rotatebox[origin=c]{90}{SMD}} 
    & DA & 0.68\textpm{0.02} & 0.68\textpm{0.02} &0.68\textpm{0.02}& 0.70\textpm{0.01}  \\
       & OA & 0.56\textpm{0.02} & 0.56\textpm{0.02} &0.56\textpm{0.02}& 0.65\textpm{0.01}  \\
       & STAS    & \textbf{0.70 \textpm{0.02}} &  \textbf{0.70 \textpm{0.02}} &  \textbf{0.70 \textpm{0.02}} & \textbf{0.75 \textpm{0.01}} \\
       & STAS/SFAS   & \textbf{0.73 \textpm{0.01}} & \textbf{0.71 \textpm{0.01}} & \textbf{0.74 \textpm{0.01}} & \textbf{0.78 \textpm{0.01}} \\
     \midrule
    \multirow{3}{*}{\rotatebox[origin=c]{90}{WVS}} 
       & OA & 0.41\textpm{0.05} & 0.41\textpm{0.05}&0.41\textpm{0.05}& 0.58 \textpm{0.05}  \\
       & STAS    & \textbf{0.84 \textpm{0.02}} &  \textbf{0.84 \textpm{0.02}} &  \textbf{0.84 \textpm{0.02}} & \textbf{0.88 \textpm{0.02}} \\
       & STAS/SFAS   & \textbf{0.86 \textpm{0.02}} & \textbf{0.87 \textpm{0.02}} & \textbf{0.85 \textpm{0.02}} & \textbf{0.90 \textpm{0.02}} \\
    \bottomrule
     \end{tabular}
    }
\end{table}
By permitting a slight time delay in decision-making, window-based localization becomes feasible. To maintain practicality, we set the \textit{look-ahead} window size, \( w_2 \), to a small value, while the \textit{look-back} \( w_1 \) is gradually increased as more data becomes available. We perform evaluations for 5 window lengths (0\%, 25\%, 50\%, 75\%, 100\%), representing percentages of the total anomalous segment lengths. As shown in Figure \ref{fig:SegmentLoc}, our model demonstrates enhanced performance in this approach, consistently improving localization accuracy across all window lengths. For window sizes between 25\% and 75\%, the maximum improvement over the best-performing method is 14\%, 58\%, and 59\% for the SMD, ASD, and WVS datasets, respectively. The final step in the localization decision process is segment localization, where the decision is made at the last consecutive time step identified as anomalous. Specifically, improvements in F1 and AUC range from 22\% to 35\% (with similar values observed in Precision and Recall). For the IPS metric, improvements range from 9\% to 45\%, demonstrating the effectiveness of our approach.
\begin{table}
    \caption{Evaluation of Segment-based Localization}
    \label{tab:SegmentLoc}
    
    \centering
    \setlength{\tabcolsep}{4pt}  
    \renewcommand{\arraystretch}{1.05}  
    \begin{tabular}{>{\raggedright\arraybackslash}p{0.05\textwidth}|c|c|c|c}  
    \toprule
         \multicolumn{1}{l|}{\textbf{Dataset}} & \textbf{Model} & \textbf{F-1} & \textbf{AUC} & \textbf{IPS} \\
    \midrule
    \multirow{4}{*}{\rotatebox[origin=c]{90}{ASD}}
       & OA  & 0.53\textpm{0.02} & 0.30\textpm{0.02} & 0.52\textpm{0.02} \\
       & DA  & 0.60\textpm{0.02} & 0.41\textpm{0.02} & 0.56\textpm{0.02} \\
       & IF  & 0.79\textpm{0.02} & 0.72\textpm{0.02} & 0.85\textpm{0.02} \\
       & STAS/SFAS    & \textbf{0.97\textpm{0.01}} & \textbf{0.95\textpm{0.01}} & \textbf{0.93\textpm{0.01}} \\
    \midrule
    \multirow{4}{*}{\rotatebox[origin=c]{90}{SMD}} 
       & OA & 0.72\textpm{0.03} & 0.78\textpm{0.02} & 0.56\textpm{0.06}  \\
       & DA  & 0.75\textpm{0.02} & 0.78\textpm{0.01} & 0.55\textpm{0.04}  \\
       & IF  & 0.63\textpm{0.02} & 0.71\textpm{0.01} & 0.59\textpm{0.04}  \\
       & STAS/SFAS  & \textbf{0.84\textpm{0.02}} & \textbf{0.87\textpm{0.01}} & \textbf{0.86\textpm{0.02}}  \\
    \midrule
    \multirow{3}{*}{\rotatebox[origin=c]{90}{WVS}} 
       & OA  & 0.57\textpm{0.02} & 0.68\textpm{0.02} & 0.57\textpm{0.05}  \\
       & IF & 0.67\textpm{0.02} & 0.76\textpm{0.02} & 0.67\textpm{0.05}  \\
       & STAS/SFAS   & \textbf{0.91\textpm{0.01}} & \textbf{0.94\textpm{0.01}} & \textbf{0.93\textpm{0.01}} \\
    \bottomrule
    \end{tabular}
\end{table}

\section{Discussion and Conclusion}
\label{sec:Conlusions}
Anomaly localization in MTS is a critical yet underexplored task. We propose a transformer-based approach that integrates transformer encoding with the novel Space-Time Anomaly Score (STAS), which is inspired by the relationship between transformer latent representations and space-time statistical models. Additionally, we enhance localization performance with the Statistical Feature Anomaly Score (SFAS) as a correction term. Our method demonstrates superior performance across various metrics and excels in the time-step, window-based, and segment-based localization tasks introduced in this study. The suggested methodology has some limitations
some of which will be considered in future work. More precisely, the STAS metric assumes that correlation coefficients remain constant over time, which may not reflect real-world dynamics. In addition, the effectiveness of window-based localization depends on the selected window length, requiring further analysis of its impact.\\

Future work will focus on enhancing STAS robustness through adaptive modules that learn time-varying correlations. We will also study window-based localization parameters, assess scalability through complexity analysis, and evaluate performance on additional datasets from diverse domains while refining the transformer encoding process for improved anomaly localization. We cordially thank for reviewers for several constructive comments.


\bibliographystyle{IEEEtran}
\bibliography{Camera_ready}

\end{document}